\documentclass[preprint,12pt]{elsarticle}

\usepackage[utf8]{inputenc}

\usepackage[usenames,dvipsnames]{xcolor}

\usepackage[normalem]{ulem}
\usepackage{booktabs}
\usepackage{amsmath,amssymb}   
\usepackage{textcomp}
\usepackage{url}

\usepackage{hyperref}

\journal{Computer Methods and Programs in Biomedicine}

\begin{document}

\begin{frontmatter}


\begin{titlepage}
    \centering
    \raggedright
    \textbf{Title Page} \\[1em]
    
     Beyond the First Read: AI-Assisted Perceptual Error Detection in Chest Radiography Accounting for Interobserver Variability \\[2em]

    \textbf{Adhrith Vutukuri\textsuperscript{*}} \\
    High School Student \\
    Tesla STEM High School \\
    \texttt{adhrith2009@gmail.com} \\[1em]

    \textbf{Akash Awasthi\textsuperscript{*}} \\
    PhD Candidate \\
    Department of Electrical and Computer Engineering \\
    University of Houston \\
    \texttt{akashcseklu123@gmail.com} \\[1em]

    \textbf{David Yang} \\
    Undergraduate Student \\
    Department of Computer Science \\
    Emory University \\
    \texttt{dongjun.yang@emory.edu} \\[1em]

    \textbf{Carol C. Wu, MD} \\
    Professor \\
    Department of Thoracic Imaging, Division of Diagnostic Imaging \\
    The University of Texas MD Anderson Cancer Center, Houston, TX \\
    \texttt{CCWu1@mdanderson.org} \\[1em]

    \textbf{Hien Van Nguyen, Ph.D.} \\
    Associate Professor \\
    Department of Electrical and Computer Engineering \\
    University of Houston \\
    \texttt{hvnguy35@central.uh.edu} \\[2em]

    \noindent\textbf{Corresponding Author:} \\
    Akash Awasthi \\
    Department of Electrical and Computer Engineering \\
    University of Houston \\
    Room N368, Cullen College of Engineering Building 1 \\
    4222 Martin Luther King Blvd, Houston, TX 77204 \\
    \texttt{akashcseklu123@gmail.com} \\[2em]

    \noindent\textit{*These authors contributed equally to this work.}

\end{titlepage}



\begin{abstract}
\textbf{Background and Objective:} Chest radiography is widely used in diagnostic imaging. However, perceptual errors—especially overlooked but visible abnormalities—remain common and clinically significant. Current workflows and AI systems provide limited support for detecting such errors after interpretation and often lack meaningful human–AI collaboration.

\noindent{\textbf{Methods:}} We introduce RADAR (Radiologist-AI Diagnostic Assistance and Review), a post‑interpretation companion system. RADAR ingests finalized radiologist annotations and CXR images, then performs regional-level analysis to detect and refer potentially missed abnormal regions. The system supports a “second-look” workflow and offers suggested regions of interest (ROIs) rather than fixed labels to accommodate interobserver variation. We evaluated RADAR on a simulated perceptual‑error dataset derived from de‑identified CXR cases, using F1 score and Intersection over Union (IoU) as primary metrics.

\noindent{\textbf{Results:}} RADAR achieved a recall of 0.78, precision of 0.44, and an F1 score of 0.56 in detecting missed abnormalities in the simulated perceptual-error dataset. Although precision is moderate, this reduces over-reliance on AI by encouraging radiologist oversight in human–AI collaboration. The median IoU was 0.78, with over 90\% of referrals exceeding 0.5 IoU, indicating accurate regional localization.

\noindent{\textbf{Conclusions:}} RADAR effectively complements radiologist judgment, providing valuable post‑read support for perceptual error detection in CXR interpretation. Its flexible ROI suggestions and non‑intrusive integration position it as a promising tool in real‑world radiology workflows. To facilitate reproducibility and further evaluation, we release a fully open‑source web implementation alongside a simulated error dataset. All code, data, demonstration videos, and the application are publicly available at: \href{https://github.com/avutukuri01/RADAR/blob/main/}{Github link}
\end{abstract}


\end{frontmatter}




\section{Introduction}
\label{sec1}
Interpretative errors in radiology continue to pose a significant challenge in clinical practice, often resulting in delayed diagnoses, inappropriate treatments, and increased patient morbidity~\cite{bruno2015,itri2018}. In CXR scans, one of the most commonly performed imaging modalities worldwide, studies have shown that radiologists fail to detect approximately 33\% of significant abnormalities in cases with confirmed disease~\cite{garland1949}. Alarmingly, the rate of missed findings in CXR interpretation has remained relatively unchanged over several decades~\cite{bruno2015,gefter2023,berlin2014}.

Among the different types of diagnostic errors, \textbf{perceptual errors}— failures to visually detect an abnormality that is present in the image—are the most prevalent, accounting for 60-80\% of interpretive mistakes~\cite{waite2016}. These errors can arise from various factors, including low lesion conspicuity, overlapping anatomy, technical limitations, incomplete search strategies (e.g., satisfaction of search), inattentional blindness, fatigue, distractions, or rushed interpretations~\cite{gefter2023, pesapane2024}. Perceptual errors are particularly difficult to address as they occur when the knowledge to make the correct diagnosis exists but the finding is not seen.

\begin{figure}[htbp]       
  \centering            
  \includegraphics[width=1\textwidth]{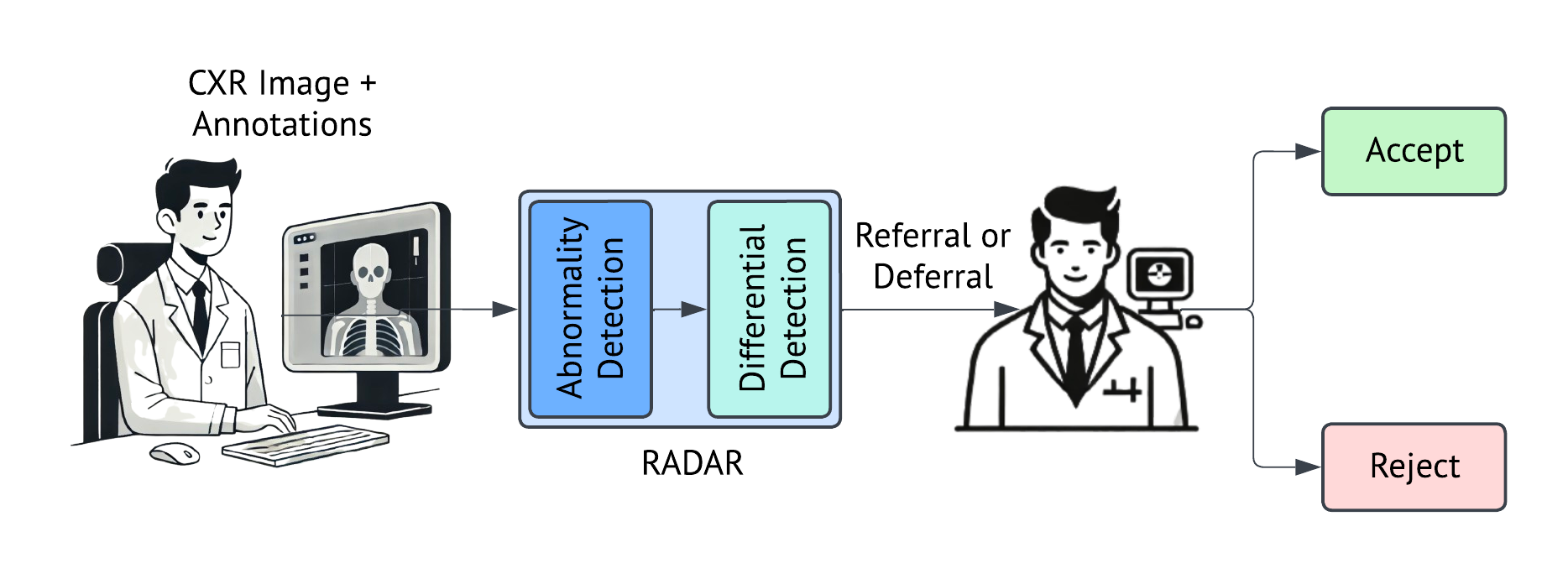}
  \caption{Overview of our proposed system, RADAR. The system receives a CXR image and the radiologist's initial annotations and then flags potential missed abnormalities. The system’s “referral” or “deferral” recommendations are finally presented back to the radiologist, who may accept or reject each flagged region.}
  \label{fig:overview}
\end{figure}

Although AI has demonstrated potential in improving diagnostic performance, most AI-based tools operate as independent detectors and provide limited assistance in identifying missed findings after a radiologist has finalized their interpretation \cite{adler2021}. Furthermore, such systems often function without regard for radiologist interaction, which can undermine trust, reduce clinical adoption, and lead to over-reliance on automation~\cite{reverberi2022}. An additional challenge arises from \textbf{interobserver variability}, especially in the labeling of abnormalities in CXR interpretation \cite{gefter2023}. This variability makes it difficult to design AI systems that generalize effectively across radiologists with different levels of experience and interpretive styles. To address these limitations, we propose \textbf{RADAR} (Radiologist-AI Diagnostic Assistance and Review), a novel lightweight post-interpretation companion system designed to assist radiologists in identifying and correcting perceptual errors in CXR interpretation. RADAR analyzes finalized radiologist annotations and image data to detect potentially unmarked abnormalities and suggest suspicious regions that may have been overlooked. Rather than replacing human judgment, RADAR promotes human–AI collaboration, functioning as a second set of eyes to support radiologists during the review phase. As illustrated in Figure ~\ref{fig:overview}, RADAR operates after the initial interpretation, taking radiologist annotations as input and generating region-level referrals—areas flagged as potentially missed but not definitively erroneous. This enables the radiologist to accept, reject, or ignore suggestions while maintaining full diagnostic control. This referral-based strategy accommodates interobserver variation and preserves clinical autonomy without enforcing deterministic abnormality labels. The main contributions of our work are as follows:

\begin{enumerate}
    \item \textbf{Clinically integrated system:} We develop RADAR, one of the first complete web-based applications for detecting and correcting visual misses in CXR interpretation, designed to work alongside radiologists.
    
    \item \textbf{Region-level referral strategy:} Our system introduces a referral-based mechanism to handle interobserver variability by identifying regions that warrant additional review, without enforcing deterministic labels.
    
    \item \textbf{Open-source tool and dataset:} We release the RADAR application as an open-source platform, along with a novel simulated error dataset containing realistic visual misses in CXR, to facilitate reproducible research in perceptual error correction.
\end{enumerate}

\section{Related Works} 
\label{relatedworks}
The challenge of perceptual errors in radiology has been extensively documented, with chest radiography (CXR) serving as a critical case study due to its diagnostic complexity and high clinical volume. Seminal work by Kundel et al. (1978) \cite{kundel1978} demonstrated that a majority of diagnostic failures arise not from flawed reasoning, but from perceptual oversights—failures to detect visible abnormalities. Follow-up studies have consistently found that even expert radiologists miss 30–40\% of clinically significant findings in CXRs with known pathology \cite{garland1949}. Despite advances in imaging and training, this error rate has remained strikingly persistent \cite{bruno2015,gefter2023,berlin2014}, indicating a need for more effective mitigation strategies.

Traditional interventions include double-reading, where a second radiologist reviews the study, and computer-aided detection (CAD) systems. While double-reading improves sensitivity, it is logistically burdensome and often impractical \cite{wagner2025}. Early CAD approaches, typically rule-based or classical machine learning, flagged potential abnormalities independently of the radiologist’s workflow. These systems were hampered by high false-positive rates and alert fatigue, limiting clinical integration \cite{vanginneken2017}. Moreover, they did not support retrospective review—leaving post-interpretation perceptual errors unaddressed.

The advent of deep learning, particularly convolutional neural networks (CNNs), has significantly advanced image classification and abnormality detection in CXR, achieving expert-level performance in identifying pathologies such as pneumothorax and lung nodules \cite{anderson2024}. However, most AI models remain stand-alone diagnostic engines, offering binary predictions without context or interaction. Their rigidity can encourage over-reliance, erode clinical trust, and—critically—fail to accommodate interobserver variability, the often-substantial differences in interpretation across radiologists \cite{quinn2023}. This variability, especially prevalent in subtle or ambiguous findings, complicates the use of fixed labels as ground truth and challenges the generalizability of AI tools.

To promote human–AI collaboration, newer systems have explored concurrent assistance paradigms. For example, Mobiny et al.(2019) \cite{mobiny2019risk} and Lakhani et al.(2020) \cite{lakhani2020} showed that AI could enhance radiologist performance when used as a real-time aid and referring to dermatologist in low confidence output. Attention-guided and uncertainty-aware models \cite{tang2018,ghoshal2020} have further improved interpretability. However, these systems still emphasize pre-interpretation support, rather than assisting radiologists in identifying missed findings after an initial assessment—precisely where perceptual errors have the greatest impact. Recent work has begun to explore second-read or review-phase AI interventions, where the system highlights potentially overlooked regions without overriding clinician judgment \cite{topff2024}. For example, CoRaX, recently proposed by Akash et al.(2024) \cite{awasthi2024}, leverages radiologist eye-gaze data to predict missed regions and associated abnormalities in chest radiographs. While this system provides valuable insights into perceptual behavior, it is sensitive to interobserver variability, as the underlying gaze and abnormality labels (Pneumonia, Atelectasis, and Edema) often reflect substantial disagreement among radiologists \cite{irvin2019chexpertlargechestradiograph,warren2018,albaum1996}. Furthermore, like many post-hoc AI tools, CoRaX \cite{awasthi2024} relies on fixed ground-truth labels during training and evaluation, which may not fully capture the nuanced spectrum of diagnostic interpretations. As a result, these systems may struggle to generalize across readers with varying experience levels or interpretive styles.

Additionally, progress in perceptual error correction is limited by a lack of purpose-built datasets. While large public CXR datasets such as CheXpert \cite{irvin2019} and MIMIC-CXR \cite{johnson2019} have advanced classification research, they do not simulate perceptual misses or enable systematic evaluation of error-detection tools.

Our work addresses these limitations on two fronts: (1) by introducing RADAR, an open-source post-interpretation AI system that generates region-level referrals to flag plausible visual misses without imposing deterministic labels, and (2) by releasing a novel simulated error dataset designed to reflect realistic interpretive oversights. Together, these contributions offer a new paradigm for radiologist–AI collaboration centered on perceptual error correction, autonomy, and reproducibility.

\begin{figure}[ht]       
  \centering            
  \includegraphics[width=1\textwidth]{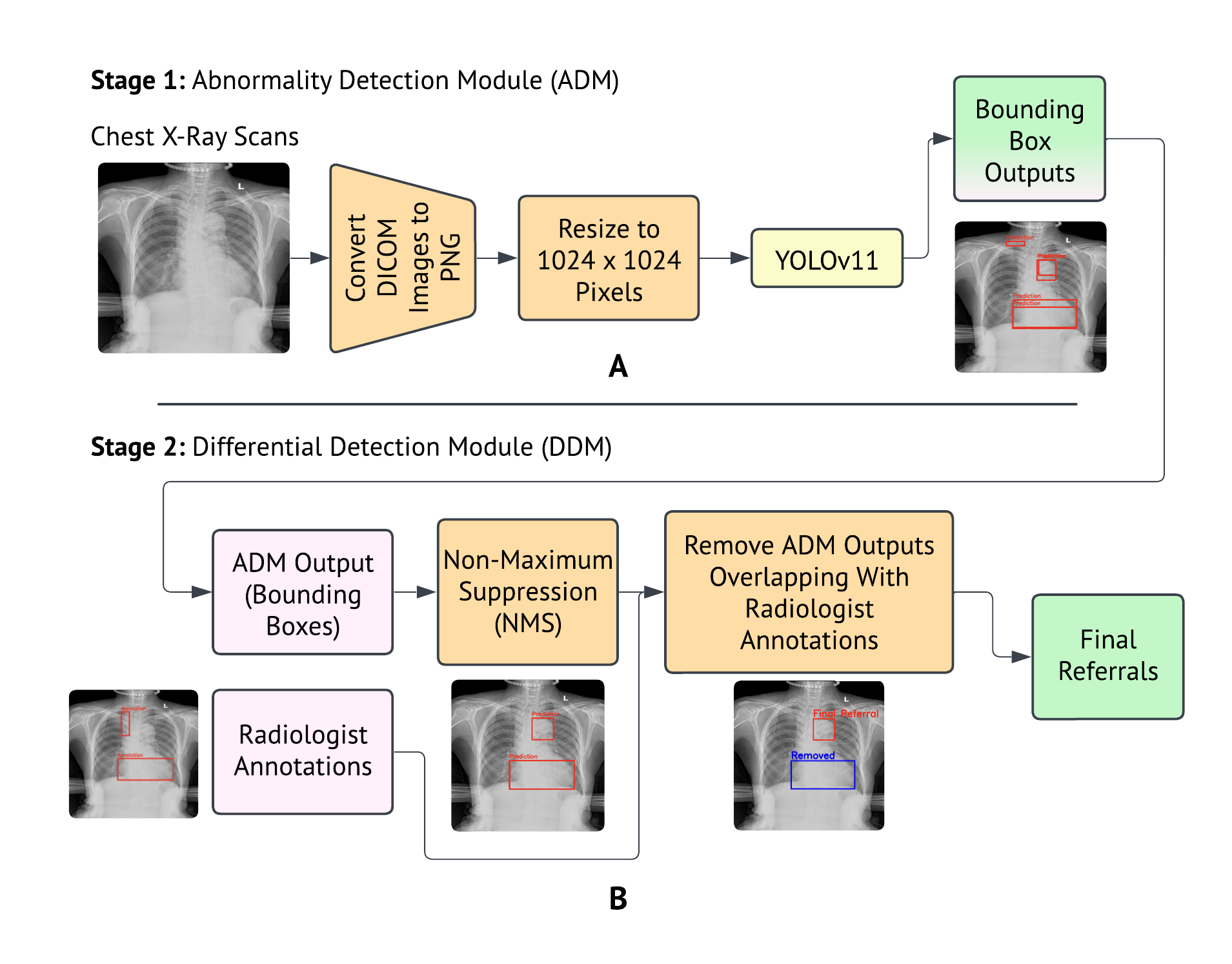}
  \caption{Overview of RADAR's workflow: the Abnormality Detection Module (ADM) proposes candidate findings which the Differential Detection Module (DDM) then compares against the radiologist's annotations to flag potential missed abnormalities.}
  \label{fig:methodology}
\end{figure}

\section{Methods}
\label{sec3}

As illustrated in Figure \ref{fig:methodology}, RADAR consists of two main components: (1) the Abnormality Detection Module (ADM) and (2) the Differential Detection Module (DDM). The ADM employs an object detection model to generate an initial set of bounding boxes that localize abnormal regions in each CXR image. DDM then compares these predicted regions with the radiologist’s original annotations to identify potential perceptual misses—abnormalities that may have been overlooked during the initial interpretation. The primary goal of this system is to highlight abnormal regions potentially overlooked by radiologists, in contrast to other post-interpretation systems for perceptual error correction, which often suffer from interobserver variability \cite{awasthi2024}.

\subsection{Abnormality Detection Module}
\label{subsec3}

The ADM is a pivotal component of RADAR and is tasked with identifying abnormal regions in CXR images and then labeling these regions with bounding boxes. As illustrated in Figure \ref{fig:methodology}A, the ADM receives CXR scans provided either as native DICOM files or as pre-compressed image formats (e.g., PNG or JPEG); all inputs are internally converted to a 1024 × 1024-pixel RGB representation before inference. Next, these images are given as input to a fine-tuned YOLOv11 object detection model, which will output bounding boxes around regions identified as abnormal. Unlike traditional two-stage detection processes, YOLO frames object detection as a regression problem and employs a unified convolutional neural network to simultaneously predict bounding boxes and class probabilities for an entire image \cite{redmon2016}. This streamlines the detection pipeline and reduces latency; both factors are essential for real-time medical applications like RADAR.

\subsection{Differential Detection Module}
\label{subsec3-2}

As illustrated in Figure~\ref{fig:methodology}B, DDM receives two inputs: (1) the bounding boxes predicted by the ADM, and (2) the bounding boxes provided by the radiologist during their initial interpretation. The DDM’s primary objective is to identify candidate regions that may have been perceptually missed by the radiologist.

 To enhance the detection of distinct missed findings, ADM-generated bounding boxes are first post-processed using non-maximum suppression (NMS) with an Intersection over Union (IoU) threshold set to zero. This configuration ensures that only one bounding box is retained among overlapping boxes (IoU $>$ 0), specifically the one with the highest confidence score, while all non-overlapping boxes are preserved. This helps eliminate redundancy without discarding potentially distinct abnormal regions. Next, the DDM performs a pairwise comparison between ADM predictions and radiologist annotations using the IoU metric. Formally, let:

\[
R = \{\,r_1,\dots,r_m\}, \quad
D = \{\,d_1,\dots,d_n\}
\]

represent the sets of bounding boxes from the radiologist and the ADM, respectively. For each ADM box \(d_i\) and each radiologist box \(r_j\), we compute:

\begin{equation}
\mathrm{IoU}(d_i, r_j)
=
\frac{\lvert d_i \cap r_j\rvert}
     {\lvert d_i \cup r_j\rvert}
=
\frac{\mathrm{area}(d_i \cap r_j)}
     {\mathrm{area}(d_i) + \mathrm{area}(r_j) - \mathrm{area}(d_i \cap r_j)}.
\label{eq:iou}
\end{equation}

A prediction \(d_i\) is flagged as a \textit{referral}—i.e., a potentially overlooked region—if it does not overlap with any of the radiologist’s annotations (i.e., maximum IoU = 0 with all \(r_j \in R\)). This is defined as:

\begin{equation}
F
=
\bigl\{\,d_i \in D \;\bigm|\;
\max_{r_j \in R}\mathrm{IoU}(d_i, r_j) = 0
\bigr\}.
\label{eq:referrals}
\end{equation}

These referral regions \(F\) are then returned to the radiologist as a second-look recommendation. Importantly, the system does not enforce any correction or override clinical judgment. Instead, it highlights regions that the ADM considers abnormal but were not initially annotated. This approach is designed to support radiologists in reviewing overlooked areas without undermining diagnostic autonomy.





\subsection{Dataset Description}
\label{subsec3-3}

We utilize the VinDr-CXR dataset~\cite{nguyen2022}, a publicly available collection of 18,000 postero-anterior (PA) CXR scans acquired from Hospital 108 and Hanoi Medical University Hospital in Vietnam. All images were annotated using a web-based PACS system (VinDr Lab) by a panel of 17 board-certified radiologists, each with at least 8 years of clinical experience. The training set (15,000 images) includes independent annotations by three radiologists per image, while the test set (3,000 images) was annotated through consensus among five radiologists to ensure higher diagnostic quality. In this study, we focus on the publicly released training portion comprising 15,000 CXR images, as distributed in the VinDr-CXR Kaggle challenge. Among these, 4,394 images contain at least one radiologist-annotated abnormality, while 10,606 images are labeled as normal. Each abnormality is marked with a rectangular bounding box, yielding precise localization labels suitable for training object detection models. 

We focus on 14 clinically relevant thoracic abnormality categories provided by the dataset: aortic enlargement, atelectasis, calcification, cardiom-\\egaly, consolidation, interstitial lung disease (ILD), infiltration, lung opacity, nodule/mass, other lesion, pleural effusion, pleural thickening, pneumothorax, and pulmonary fibrosis. These classes span a range of common and subtle pathologies in chest radiographs and form the basis for evaluating the performance of our ADM.

\subsection{Dataset Preprocessing}
\label{subsec3-4}

The original CXR scans are in DICOM format with varying spatial dimensions. We converted each image to PNG format and resized it to 1024$\times$1024 pixels for consistency across the training pipeline. This preprocessing step ensures compatibility with the YOLOv11 architecture used in the ADM, which expects uniformly sized inputs. Given the class imbalance between normal (10,606) and abnormal (4,394) cases, we constructed a balanced dataset to prevent model bias toward the majority class. To achieve this, we randomly selected 4,394 normal images and paired them with all 4,394 abnormal images, resulting in a total of 8,788 scans. This balanced dataset was then split into 80\% for training and validation, and 20\% for held-out testing. The training/validation split was further divided into 80\% training and 20\% validation subsets.

\subsection{Error Dataset}
\label{subsec3-5}

\begin{figure}[ht]      
  \centering           
  \includegraphics[width=1\textwidth]{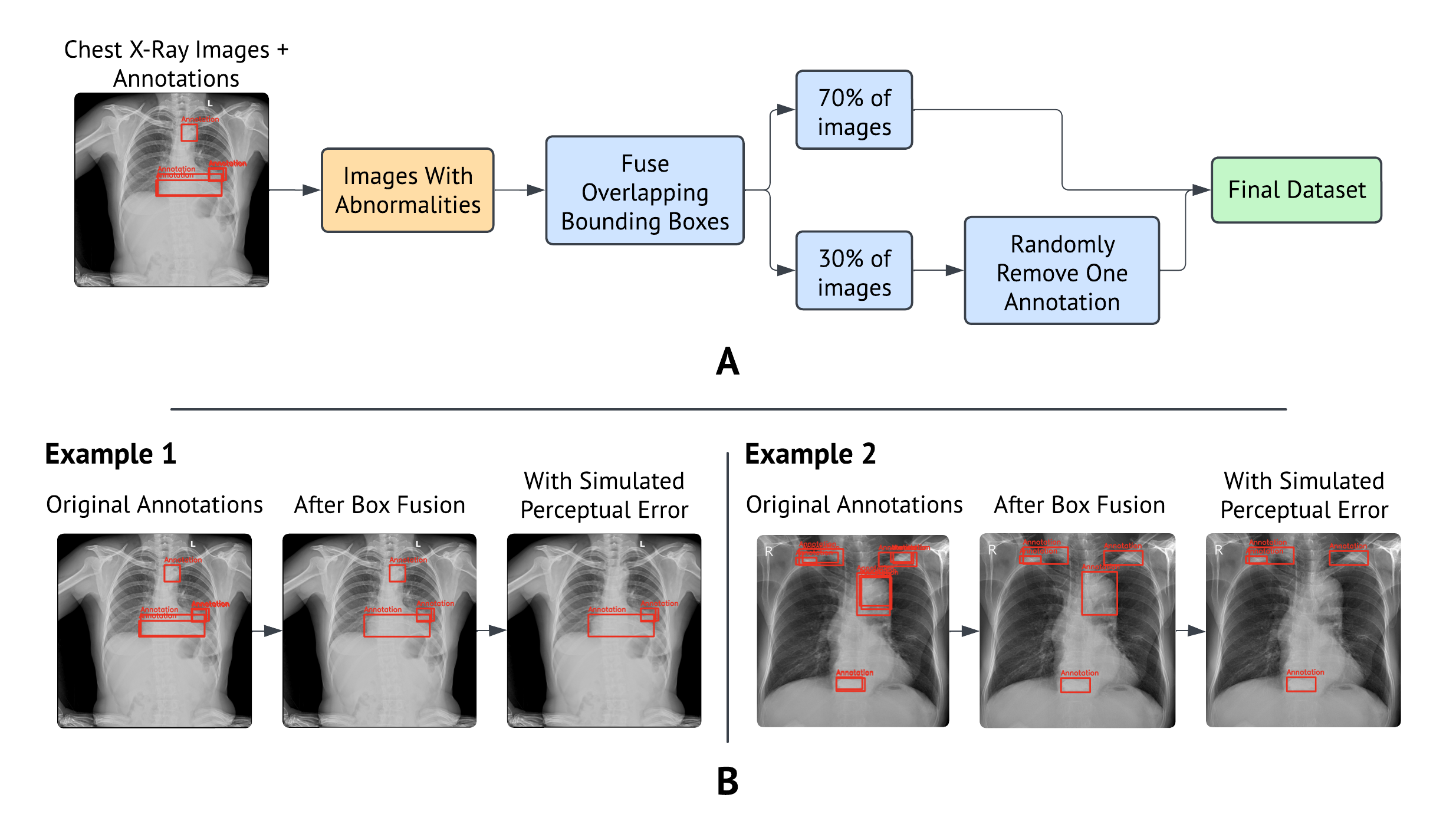}
  \caption{Representation of the proposed data-alteration pipeline for simulating perceptual errors. Subfigure (\textbf{A}) illustrates the data-alteration pipeline: abnormal chest X-rays first undergo IoU-based fusion of overlapping bounding boxes; then, in 30\% of images, one fused annotation is randomly removed to simulate perceptual errors, producing the final dataset. Subfigure (\textbf{B}) shows two example cases, each with original overlapping radiologist annotations (left), fused boxes after merging at an IoU threshold of 0.3 (center), and the corresponding simulated-error image with one annotation removed (right).}
  \label{fig:error-pipeline}
\end{figure}

Because no publicly available dataset specifically contains radiologist perceptual errors, we constructed a simulated error dataset to test RADAR’s post-interpretation capabilities (Figure~\ref{fig:error-pipeline}). We began with the aforementioned test set, which contains multiple overlapping bounding boxes due to independent annotations from multiple radiologists. These overlaps were resolved using bounding box fusion: we applied a greedy merging algorithm to fuse all bounding boxes with an IoU $\geq$ 0.3, producing a single representative bounding box per lesion. To simulate visual misses, we randomly removed one of the fused abnormality boxes in approximately 30\% of the abnormal cases. This simulates a common type of perceptual error in clinical practice, where a radiologist may fail to detect one of several abnormalities due to limitations in attention, visual search, recognition, and satisfaction of search. 

The resulting test set includes both complete and altered annotations, allowing us to assess RADAR's ability to identify missed abnormalities in a controlled but clinically plausible setting. By introducing structured noise, this approach serves as a proxy for real-world perceptual oversights.

\subsection{Experimentation}
\label{sec4}

This section outlines the experimental design used to train and evaluate the ADM and the complete RADAR system. We first describe the training and validation protocols for ADM, followed by the framework used to assess RADAR’s performance in detecting missed abnormalities.

\subsubsection{ADM Training and Evaluation Protocol}
\label{subsec4-1}

To identify the most effective detection paradigm for CXR abnormality localization, we trained two configurations of the ADM using the YOLOv11 object detection architecture: (1) a multi-class model, which predicts individual types of thoracic abnormalities; and (2) a single-class model, which collapses all abnormalities into a single “any abnormality” class. This comparison was designed to assess whether fine-grained or generalized detection is more robust for downstream perceptual error correction.

In both configurations, we fine-tuned YOLOv11 using default augmentation strategies aimed at improving generalization on CXR data. These included hue, saturation, and value jittering; horizontal flipping with a probability of 0.5; random translation (±10\%); scaling (±50\%); and mosaic augmentation.

Training was conducted using the AdamW optimizer with an initial learning rate of $0.002$, a momentum of $0.9$, and a weight decay of $5 \times 10^{-4}$. A linear warm-up schedule was used over the first three epochs, during which the momentum ramped up from $0.8$, and the bias learning rate was set to $0.1$. After warm-up, the learning rate decayed linearly from $0.002$ to $2 \times 10^{-5}$. Training was performed for over 100 epochs with a batch size of 8, constrained by the memory of an NVIDIA L4 GPU. Model performance was evaluated on a held-out validation set using standard object detection metrics: mean average precision at an IoU threshold of 0.5 (mAP50), precision, and recall. The model that achieved the best balance between sensitivity and specificity was selected as the final ADM architecture.
\subsubsection{RADAR Evaluation Framework}
\label{subsec4-2}

To evaluate RADAR’s ability to detect perceptual errors—i.e., abnormalities missed by radiologists—we used the simulated error dataset described in Section~\ref{subsec3-5}. The goal was to assess how effectively RADAR identifies and flags these visual misses through its referral mechanism.

We introduce four outcome categories to characterize the performance of the DDM:

\begin{itemize}
    \item \textbf{True Referral (TR)}: RADAR correctly identifies a missed abnormality and presents it as a referral. This corresponds to a clinically meaningful detection that a radiologist would accept upon review.
    
    \item \textbf{False Referral (FR)}: RADAR incorrectly flags a normal region as abnormal. This represents a false positive referral that would likely be rejected by a radiologist.
    
    \item \textbf{False Deferral (FD)}: RADAR fails to identify a region that was missed by the radiologist. This is a critical false negative scenario where RADAR offers no referral despite a real oversight.
    
    \item \textbf{True Deferral (TD)}: RADAR correctly refrains from making a referral, either because no abnormality is present or because the radiologist has already identified all existing abnormalities.
\end{itemize}

Using these outcome categories, we computed the Precision, Recall, F1-score and Accuracy.

\section{Results and Discussion}
\label{sec5}

In this section, we first quantify the ADM's performance in localizing abnormal regions (Section \ref{subsec5}) and then analyze the effectiveness of RADAR in correcting perceptual errors (Section \ref{subsec5-2}).

\subsection{ADM Evaluation}
\label{subsec5}
We evaluated two configurations of the ADM in RADAR using the YOL-\\Ov11 object detection model: a single-class setup and a multi-class setup. As shown in Table \ref{tab:YOLO-model-results}, the single-class setup outperformed the multi-class variant across all metrics—precision (0.561 vs. 0.402), recall (0.388 vs. 0.359), and mAP@0.5 (0.433 vs. 0.330). This superior performance is likely due to reduced classification complexity and fewer inter-class confusions during training. Based on these findings, we selected the single-class model as the final ADM for RADAR.

Importantly, the single-class approach offers added robustness in the context of chest radiography, where interobserver variability among radiologists is known to be high. Prior studies have reported disagreement rates of up to 20\% in patients with radiographic findings~\cite{bruno2015, gefter2023,quekel2001, balabanova2005, young1994} A significant proportion of these disagreements arise not from whether an abnormality is present, but rather from how it is classified. By collapsing all abnormalities into a single class during training, the ADM reduces susceptibility to such variability and minimizes label noise during supervision. This design choice is particularly advantageous in CXR interpretation, where boundary definitions between categories—e.g., consolidation vs. infiltration—are often ambiguous. Furthermore, the single-class detection paradigm aligns more closely with RADAR’s role as a perceptual aid rather than a disease classifier. The system is tasked not with making fine-grained diagnostic distinctions, but with drawing the radiologist's attention to potentially overlooked regions of abnormality. In this context, sensitivity to any suspicious area is more clinically valuable than perfect classification granularity.

\begin{table}[htbp]
  \setlength{\belowcaptionskip}{8pt}
  \centering
  \caption{Comparison of performance metrics for the fine-tuned single-class and multi-class YOLO models.}
  \label{tab:YOLO-model-results}
  \renewcommand{\arraystretch}{1.2} 
  \begin{tabular}{lccc}
    \toprule
    \textbf{Model} & \textbf{Precision} & \textbf{Recall} & \textbf{mAP@0.5} \\
    \midrule
    Single-Class & 0.561 & 0.388 & 0.433 \\
    Multi-Class  & 0.402 & 0.359 & 0.330 \\
    \bottomrule
  \end{tabular}
\end{table}

\subsection{RADAR Evaluation}
\label{subsec5-2}
\begin{figure}[htbp]       
  \centering            
  \includegraphics[width=1\textwidth]{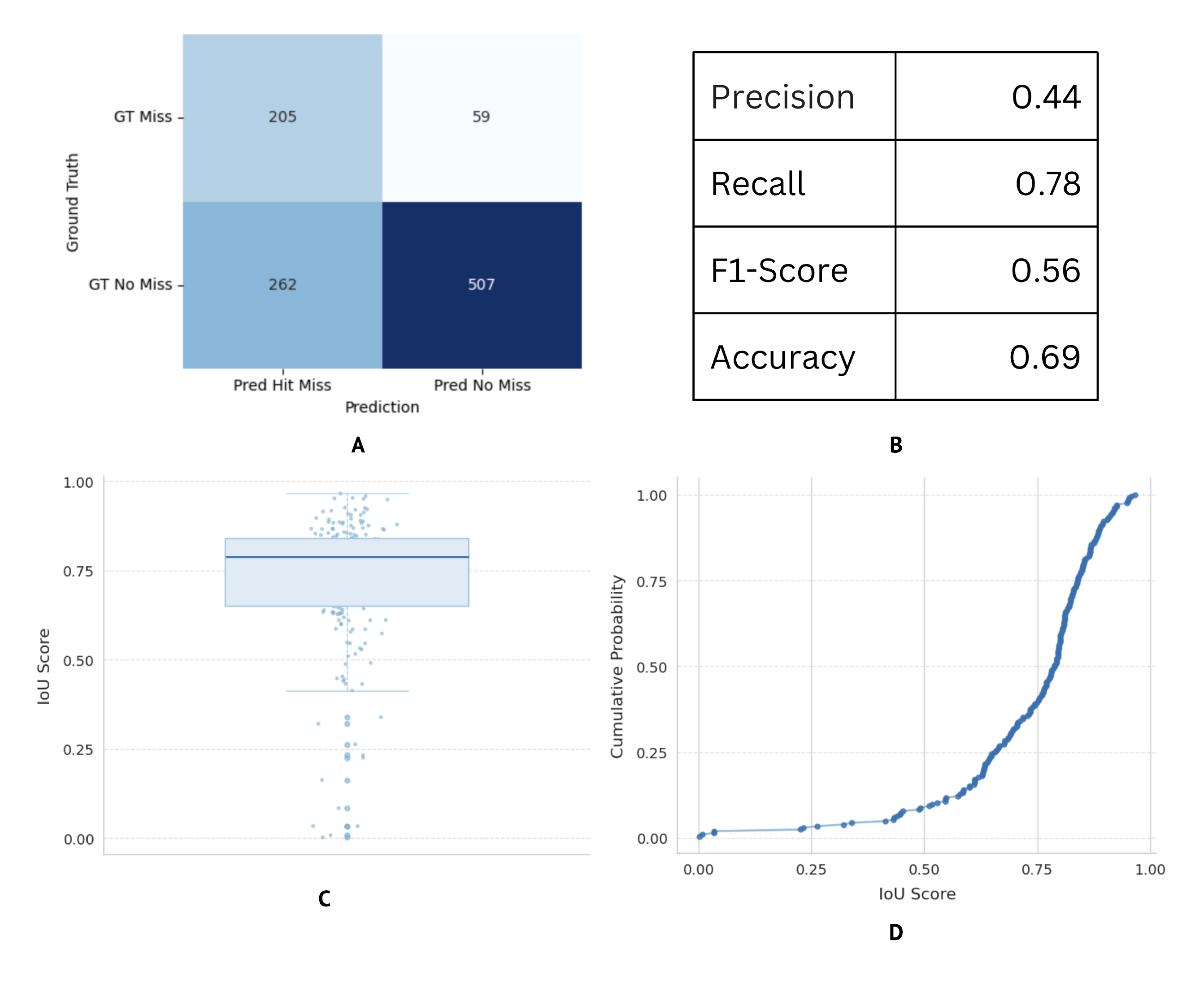}
  \caption{Visualization of RADAR perceptual error correction results. Subfigure (\textbf{A}) is a confusion matrix showing RADAR's classification of perceptual misses. Subfigure (\textbf{B}) is a table showing the precision, recall, F1-score, and accuracy of RADAR correcting perceptual errors. Subfigure (\textbf{C}) is a box and whisker plot showing the IoU score between true referrals from RADAR and the error they are correcting. Subfigure (\textbf{D}) is a cumulative distribution plot of the IoU overlap between RADAR's true referrals and ground-truth ROIs.}
  \label{fig:RADAR_Eval}
\end{figure}

Figure~\ref{fig:RADAR_Eval} presents a summary of RADAR's performance in identifying visual misses, evaluated through classification metrics and spatial localization accuracy. As shown in the confusion matrix and the performance table (Figures~\ref{fig:RADAR_Eval}A and B), RADAR achieved a recall of 0.78, indicating that it correctly identified 204 out of 263 missed abnormalities in the simulated perceptual error data set. This high recall suggests that RADAR has substantial potential to recover clinically significant findings that would otherwise go unnoticed during routine interpretation. However, this comes at the cost of some false positives, yielding a precision of 0.44. The corresponding F1-score of 0.56 represents a balanced summary of this tradeoff. The overall accuracy of 0.69 reflects RADAR’s aggregate ability to make correct referral or non-referral decisions in the presence of simulated radiologist error.

In addition to classification performance, we evaluated the spatial precision of RADAR’s true referrals using the IoU metric. As shown in Figures~\ref{fig:RADAR_Eval}C and D, the median IoU between true RADAR referrals and corresponding missed abnormalities was approximately 0.78, and most values fell in the 0.65–0.85 range. In particular, more than 90\% of referrals exceeded the 0.5 IoU, indicating a consistent and clinically reliable location of visual misses.

\begin{figure}[htbp]       
  \centering            
  \includegraphics[width=1\textwidth]{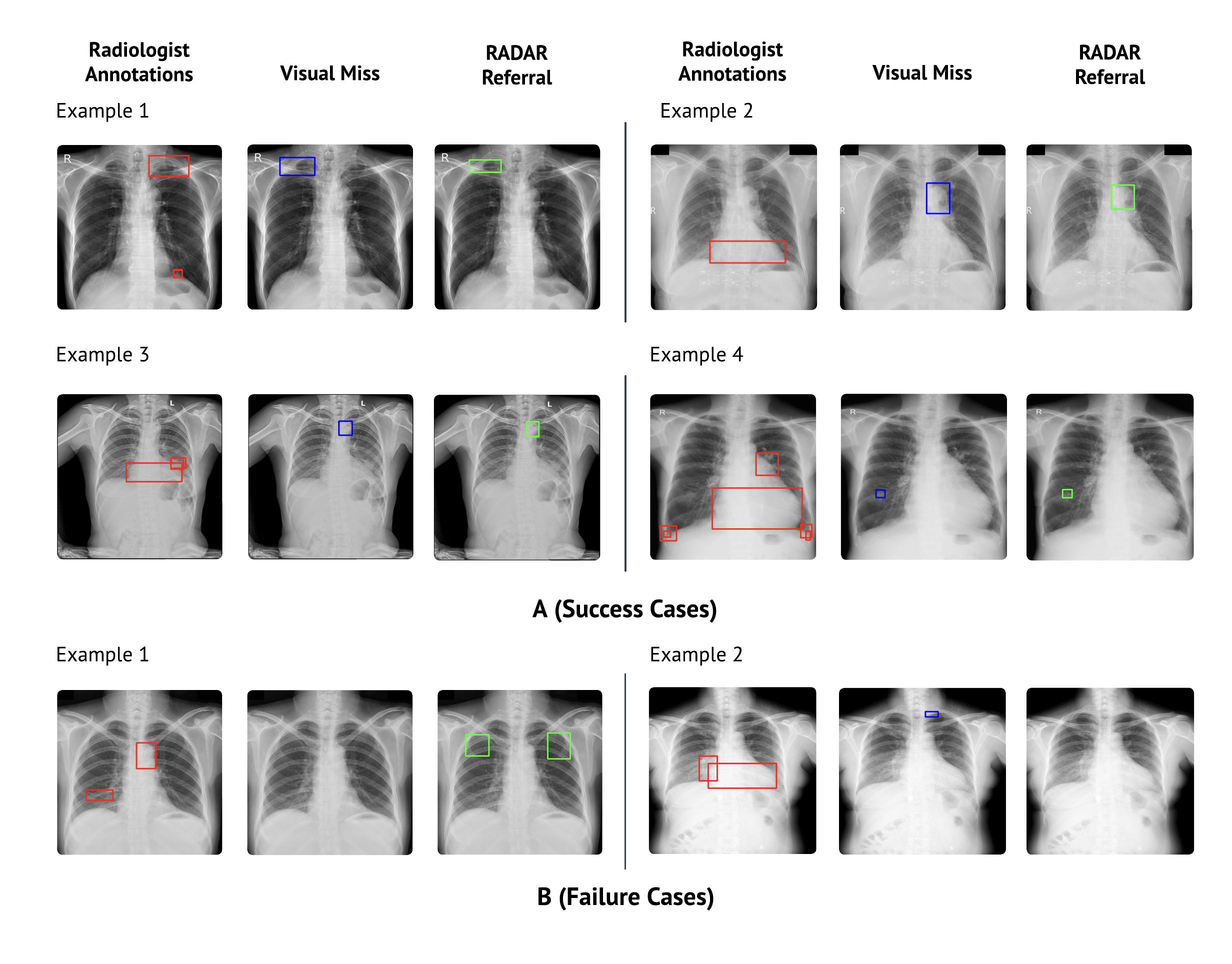}
  \caption{Visualization of six example cases showing radiologist's annotations, simulated visual misses from the simulated error dataset, and true RADAR referrals. Subfigure (\textbf{A}) shows four true referral cases where RADAR correctly identifies a visual miss. Subfigure (\textbf{B}) shows two examples where RADAR gives an incorrect output. Example 1 shows a false referral case, and Example 2 shows a false deferral case.}
  \label{fig:RADAR_Qual_Results}
\end{figure}

In addition to quantitative performance, qualitative examples further illustrate the capability of RADAR to identify perceptual misses and its limitations. Figure~\ref{fig:RADAR_Qual_Results}A presents successful referral cases that highlight clinically relevant detection patterns. In Example 1, the radiologist captured an abnormality in the left lung, while a similar but distinct abnormal region on the right, located near the tracheal shadow, was missed. RADAR correctly identified this second region, demonstrating its ability to identify bilateral abnormalities, even when only one side is annotated. This capability is clinically important, as radiologists often anchor on a single lesion and overlook contralateral counterparts. In Example 4, RADAR identified a small and subtle abnormality in the lower right lung, which was missed in the simulated radiologist interpretation. The flagged region was precisely localized with a tight bounding box, suggesting that the model is sensitive to focal, low-salience abnormalities that are prone to human oversight.

Conversely, Figure~\ref{fig:RADAR_Qual_Results}B shows examples of failure modes that reflect current limitations of RADAR. In Example 1 (false referral), the radiologist had not missed any abnormalities, yet RADAR predicted spurious bounding boxes in the upper lung fields bilaterally. Such errors could stem from overgeneralization of texture cues or imaging artifacts. In Example 2 (false deferral), the radiologist overlooked an abnormal region in the upper lobe of the left lung, yet RADAR failed to flag it for referral. This case represents a missed opportunity for corrective intervention and underscores the importance of improving RADAR’s sensitivity in subtle but clinically significant regions. These qualitative examples complement the aggregate metrics by illustrating how RADAR performs under specific clinical scenarios—particularly in detecting bilateral findings, small nodules, and ambiguous cases. 

In summary, RADAR demonstrates a potential to function as a perceptual error-correction aid for radiologists. Its high recall and spatial precision make it well-suited for flagging overlooked findings, while its moderate precision underscores the importance of maintaining the radiologist-in-the-loop to adjudicate borderline or ambiguous cases. These results provide evidence that a referral-based AI framework can meaningfully enhance interpretive accuracy in chest radiography.

\section{Web Application Framework}

\begin{figure}[htbp]       
  \centering            
  \includegraphics[width=1\textwidth]{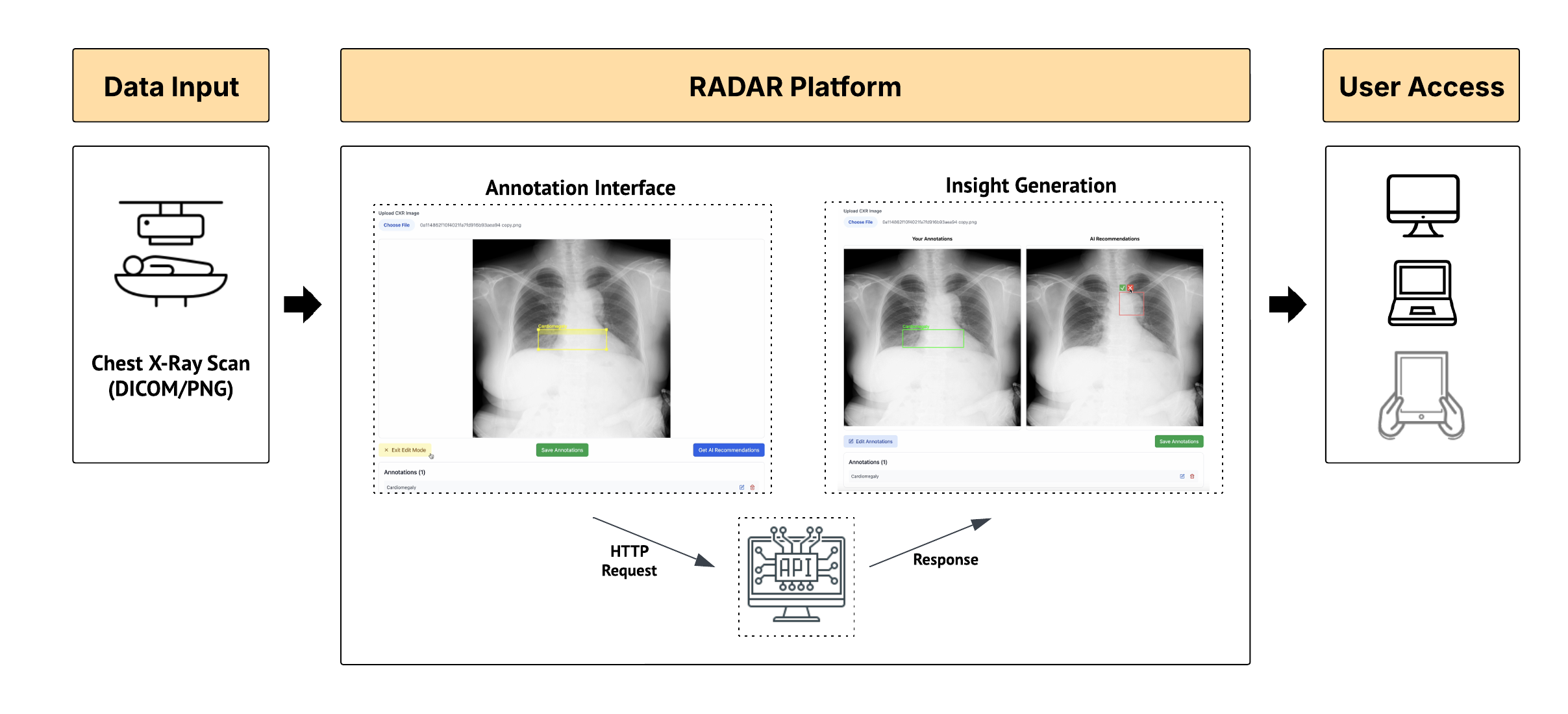}
  \caption{Architecture of RADAR platform. }
  \label{fig:RADAR_Application}
\end{figure}

After creating the RADAR flow, we implemented the system as a web application that radiologists can use intuitively to receive real-time referrals for correcting perceptual errors. As shown in Figure \ref{fig:RADAR_Application}, the application allows users to load Chest X-Ray scans and annotate them directly in the browser. Once the image is annotated, the radiologist clicks on the "Get AI recommendations" option. The image and its annotations are then sent to the RADAR back-end, which was implemented via Flask API. We exposed the local Flask API via an ngrok tunnel, providing a temporary HTTPS endpoint that allowed the online application to communicate securely with the API. The API processes the data and returns region-based referrals. The application displays these referrals in a side-by-side view, with the radiologist's annotations on the left and RADAR's referrals on the right. The radiologist may accept or reject each referral: rejecting keeps the original annotation, whereas accepting applies the recommended bounding box, which can still be adjusted and labeled with the appropriate abnormality.

\section{Computation and Response Time}
\label{sec6}
We assessed RADAR’s computational overhead and latency to ensure the system remains accessible and capable of near-real-time feedback. Benchmarking was performed on both an Apple MacBook Pro equipped with an M4 chip and on an NVIDIA L4 GPU hosted in Google Colab. On the MacBook, end-to-end inference required an average of 8 – 12 s when processing DICOM inputs, with the majority of this time attributed to DICOM decoding and pre-processing. Replacing the DICOM file with a PNG reduced total runtime to approximately 2 – 4 s, although performance fluctuated depending on concurrent local workloads. On the L4 GPU, the same DICOM workflow completed in roughly 5 s, while PNG inputs were processed in about 1 s. All Colab timings were collected inside the notebook without routing requests through ngrok, thereby isolating compute overhead from external network latency. Taken together, these benchmarks show that RADAR delivers clinically actionable results in just a few seconds even on a consumer-grade laptop, and in near-real-time on a modest cloud GPU. This low computational footprint lowers the cost-of-entry and makes advanced decision support by RADAR accessible to both large academic centers and resource-constrained clinics. 

\section{Ablation}

To assess potential improvements in referral precision, we conducted an ablation study by modifying RADAR’s pipeline to incorporate a pre-trained DenseNet-121 classifier \cite{tang2020} as an upstream gating mechanism before the detection module. The motivation for this experiment stemmed from the observation that false referrals often arise from normal radiographs. By identifying and excluding clearly normal cases early in the pipeline, we hypothesized that the overall precision of the system could be improved by avoiding unnecessary referral generation. In this classifier-gated variant, each chest radiograph is first passed through the DenseNet-121 model, which produces a binary prediction: "normal" or "abnormal." Radiographs classified as normal bypass the ADM entirely, leading to no bounding box predictions or referrals. Only abnormal cases proceed to the YOLO detector and DDM for region-level analysis and referral generation. This configuration thus serves as a conservative filtering step aimed at reducing false positives from structurally normal CXRs.

Table~\ref{tab:ablation-classifier} summarizes the results. Adding the classifier resulted in a notable increase in precision (from 0.44 to 0.48) and overall accuracy (from 0.69 to 0.71), consistent with the intended effect of reducing incorrect referrals. However, this came at the cost of reduced recall (from 0.78 to 0.69), reflecting missed opportunities where the classifier mistakenly filtered out cases that contained subtle abnormalities. The F1-score remained unchanged (0.56), indicating a net trade-off between sensitivity and specificity. High recall is critical in the context of perceptual error correction, where missing true abnormalities can have serious clinical consequences. For this reason, we chose not to include the classifier gating mechanism in the final RADAR pipeline.

\begin{table}[htbp]
  \centering
  \caption{Ablation study comparing detection performance with and without a preceding pretrained classifier. Including a classifier before YOLO improves precision and accuracy, but decreases recall.}
  \label{tab:ablation-classifier}
  \renewcommand{\arraystretch}{1.2} 
  \begin{tabular}{lcccc} 
    \toprule 
    \textbf{Model} & \textbf{Precision} & \textbf{Recall} & \textbf{F1-score} & \textbf{Accuarcy} \\ 
    \midrule
    YOLO & 0.44 & 0.78 & 0.56 & 0.69 \\
    Classifier + YOLO & 0.48 & 0.69 & 0.56 & 0.71 \\
    \bottomrule
  \end{tabular}
\end{table}

\section{Limitations and Future Work}
\label{sec7}

This study introduces RADAR, a web-based tool developed to identify visual misses in chest X-rays by providing automated referral suggestions for review. Although our results demonstrate promising performance, several limitations remain. First, the simulated perceptual error dataset does not fully capture the complexity of human perceptual behavior. Factors such as fatigue, search satisfaction, and prior expectations are not modeled. Future studies should incorporate real-world perceptual miss data collected by eye tracking, double reading,or reader disagreement to more accurately represent clinical error. Second, as a technical proof-of-concept, this work has not yet undergone clinical validation. No user studies or workflow trials have been conducted to assess how RADAR performs in practice or how radiologists respond to its referrals. Future work will focus on evaluating the impact of the tool on diagnostic precision, decision confidence, and user trust in real world settings. Third, although RADAR demonstrates high recall (0.78) and strong localization (median IoU = 0.78), its moderate precision (0.44) indicates the need to improve specificity. Future iterations will explore improved object detection architectures and gating mechanisms to reduce false positives without compromising safety. In general, this paper represents a foundational step in building clinically aware AI assistants for perceptual error mitigation. Subsequent work will focus on improving the robustness of the model and integrating RADAR into radiologist workflows for prospective evaluation.

\section{Conclusion}
\label{sec8}

We introduced RADAR, a novel AI framework for detecting perceptual errors in chest X-ray interpretation by identifying missed abnormalities and generating targeted referral suggestions. By combining a fine-tuned object detector with a differential detection module, RADAR proactively flags potentially overlooked findings to assist radiologists. Our experimental evaluation, based on a simulated perceptual miss dataset, demonstrates RADAR’s ability to recover a substantial fraction of visual misses with high spatial fidelity. As a technical contribution, this work proposes a shift from conventional passive detection systems to proactive error-aware support tools, with potential applications in reducing clinically significant oversights. While further validation is needed, particularly in real-world clinical settings, RADAR offers a promising foundation for the next generation of intelligent diagnostic assistants. Future work will refine its precision, incorporate real user behavior, and evaluate its practical impact through clinical integration and reader studies.

\section{Ethics statement}
This study did not involve direct data collection from human participants. It utilized publicly available datasets  VinDr-CXR dataset that were collected with appropriate ethical approvals and de-identified to protect participant privacy. Therefore, additional ethical approval was not required for this study.

\section{Acknowledgment}

This work was supported in part by the National Institutes of Health under Grant 1R01CA277739. The content is solely
the responsibility of the authors and does not necessarily represent the official views of the National Institutes of Health




\bibliography{radar_refs}          

\end{document}